\documentclass[letterpaper]{article} 

\usepackage[preprint]{aaai2027}  

\usepackage[hyphens]{url}  
\usepackage{graphicx} 
\usepackage{natbib}  
\usepackage{caption} 
\newcommand{\R}{\mathbb{R}}
\newcommand{\E}{\mathbb{E}}
\newcommand{\ind}{\mathbf{1}}
\usepackage{amsmath}
\usepackage{amssymb}
\usepackage{booktabs}
\usepackage{comment}

\title{Double Descent in Gradient Boosting Decision Trees via Split-Candidate Scaling}
\author{Ryuichi Kanoh$^{\mathbf{1,2}}$}
\affiliations{
    $^{1}$The University of Electro-Communications \\
    $^{2}$National Institute of Informatics\\
    ryuichi.kanoh@uec.ac.jp
}

\begin{document}

\maketitle

\begin{abstract}
\emph{Double descent} is commonly studied by scaling an explicit capacity
parameter, such as neural-network width. For \emph{gradient boosting decision trees}
(GBDTs), however, an analogous single-axis capacity parameter has not been
established. We propose \emph{the number of split candidates} as an operational
capacity parameter for GBDTs. Holding other training controls fixed, increasing
the split-candidate budget refines the feature-quantization grid and expands the
dictionary of root-to-leaf paths from which boosting selects its updates. To
analyze this expansion, we construct an empirical tree-kernel diagnostic that
summarizes how candidate-induced paths group the training examples. A regime
in which the empirical kernel rank grows toward the sample size and very small
positive eigenvalues emerge exposes noise-sensitive directions; in this regime,
test error peaks before decreasing again at larger split-candidate budgets. This
perspective predicts that deeper trees should reach the regime with fewer split
candidates, larger training sets should require finer grids, and label noise
should make the peak more pronounced.
Experiments support these predictions and show test-error peaks at intermediate
split-candidate budgets across XGBoost, LightGBM, and CatBoost, whereas a
random-forest control improves monotonically under the same split-candidate
sweep. Taken together, our analysis and experiments support split-candidate
scaling as a single-axis capacity intervention for studying GBDTs and suggest
that the observed double descent arises from an interaction between
candidate-induced geometry and boosting dynamics.
\end{abstract}

\section{Introduction}
In supervised prediction tasks, a model is trained on a finite sample but evaluated
on new, unseen examples.  As the model becomes more expressive, its training
error typically decreases because it can fit the observed sample more closely.
The central question is whether this improved fit also leads to better
generalization.  Classical intuition says that test error should first
decrease, as the model captures useful structure, and then increase, as the
model begins to fit noise. \emph{Double descent} describes a different pattern
\citep{belkin2019reconciling,nakkiran2020deep}.  Test error can peak near the
point where the model first fits the training sample, but then decrease again
as capacity continues to grow.  In other words, making a model more flexible
can hurt generalization at an intermediate scale and then help it again in a
more overparameterized regime. In neural networks, this model-wise pattern is
commonly studied by increasing network width, which corresponds to the number of parameters \citep{nakkiran2020deep}.

Although earlier work reported double descent in \emph{gradient boosting
decision trees} (GBDTs) using raw parameter count as the capacity axis
\citep{belkin2019reconciling}, the interpretation of this result remains
debated because the curve was constructed using two distinct mechanisms for
increasing that count.  It first grew by adding boosting rounds; the number of rounds was then fixed, and
independently trained boosted models were averaged to increase the count
further.  \citet{curth2023uturn} subsequently showed that the second descent
coincides with the transition between these mechanisms, suggesting that the
post-peak improvement may reflect variance reduction from ensembling rather
than increased expressivity of a single GBDT.  More generally, counts of trees,
nodes, or leaves do not by themselves define a consistent capacity
intervention: adding rounds also extends the optimization trajectory, while
increasing depth changes both interaction structure and partition geometry.
This unresolved methodological issue leaves a central question: ``Can double
descent in GBDTs arise under a single, consistent capacity intervention?''

\begin{figure}[t]
  \centering
  \includegraphics[width=\columnwidth]{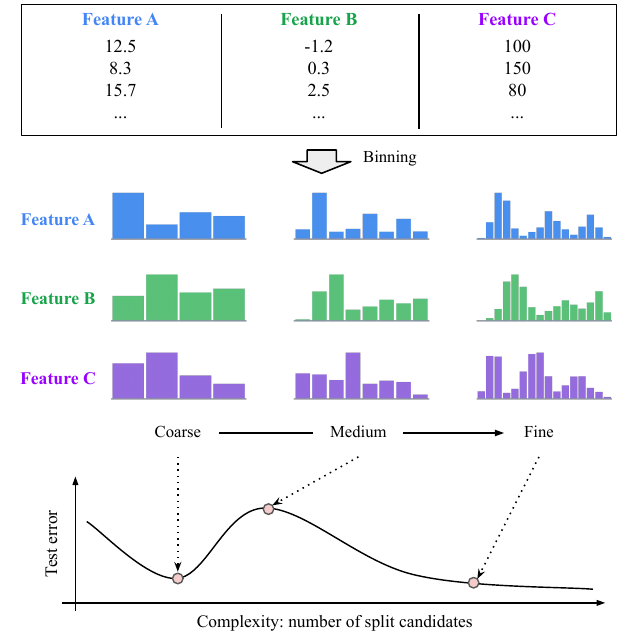}
  \caption{Schematic overview of split-candidate scaling and the associated
  double-descent mechanism.}
  \label{fig:split-candidate-schematic}
\end{figure}

This paper proposes an operational axis for addressing this question: \emph{the number of split
candidates}.  Modern GBDT systems, including
XGBoost~\citep{chen2016xgboost}, LightGBM~\citep{ke2017lightgbm}, and
CatBoost~\citep{prokhorenkova2018catboost}, first discretize continuous
features into a finite set of candidate thresholds.  A coarse grid permits
only a limited collection of leaf regions, whereas a finer grid gives the
learner a larger dictionary of root-to-leaf path features.  Thus, increasing
the split-candidate count directly enlarges the set of update directions
available to boosting.
GBDT systems expose binning controls to trade accuracy against split-search
time, memory use, and communication cost.  Split-candidate count is therefore
both a capacity measure and an implementation-level control.
Figure~\ref{fig:split-candidate-schematic} illustrates the high-level picture:
as the numerical grid is refined, the path-feature dictionary expands and can
produce a test-error peak at an intermediate grid resolution.

We explain the mechanism through three linked views.
The \emph{tree view} is the conventional description: boosting adds trees whose
leaf regions are determined by available split candidates.  The \emph{feature
view} rewrites exactly the same model as a linear combination of root-to-leaf
path indicators.  Increasing the candidate count enlarges this dictionary,
making the same tree model appear as a wider linear model with more
coefficients.  The \emph{kernel view} studies
this expansion through a reference tree kernel on the finite training sample.
As paths distinguish more training points, the kernel rank approaches the
sample size and small-eigenvalue directions can appear.  Under weak
regularization, these small-eigenvalue directions are sensitive to label noise
and can create a test-error peak.  This chain predicts that deeper trees reach
the rank transition with fewer candidates, larger training sets require finer
grids, and label noise makes the peak more visible.

Our contributions are summarized as follows:

\begin{itemize}
\item Identification of split-candidate count as a single-axis capacity
intervention for GBDTs through a three-view formulation connecting leaf
regions, path features, and empirical tree-kernel spectra.
\item Empirical tests of predicted peak shifts across depth, sample size, label
noise, tree-ensemble libraries, and benchmark datasets.
\end{itemize}

\section{Related Work}

\paragraph{Gradient boosting decision trees.}
GBDTs originate from functional gradient descent with regression trees as base
learners \citep{friedman2001greedy}.  Modern libraries such as XGBoost,
LightGBM, and CatBoost have made GBDTs a practical
workhorse for tabular prediction; they are widely used in applied systems and
machine-learning competitions, and empirical comparisons continue to find tree
ensembles strong on typical tabular data
\citep{fu2019experimental,grinsztajn2022why}.  A common implementation device
is to quantize continuous features before tree induction and then evaluate split
candidates on the resulting bins. GBDT libraries expose binning controls
primarily to manage the computational cost of split search. In this study, we
use the same controls to vary the number of available split candidates.
The way trees are constructed varies substantially across systems and variants:
libraries differ in their growth rules and tree shapes, and some methods inject
randomness through candidate sampling or noisy split scores rather than using
only deterministic greedy splits \citep{geurts2006extremely}.
\citet{ustimenko2023gradient} further develop a fixed tree-kernel perspective
for gradient boosting with symmetric decision trees and relate the resulting
dynamics to Gaussian-process inference.  Building on this viewpoint, we track
how the induced empirical spectrum changes as the candidate set produced by
binning is varied, with the goal of studying split-candidate scaling as a
capacity axis in widely used GBDT implementations.

\paragraph{Double descent.}
The double-descent risk curve was popularized as a way to reconcile classical
statistical intuition with high-capacity models that interpolate training data
\citep{belkin2019reconciling}.  Precise analyses now exist for linear
ridgeless regression, including \citet{hastie2022surprises},
\citet{belkin2020twomodels}, and \citet{holzmueller2021universality}, and for
random-feature regression, whose feature construction originates from
\citet{rahimi2007random} and whose double-descent asymptotics are analyzed by
\citet{mei2022generalization}.  These analyses emphasize the role of the
empirical feature spectrum near the point
where the model first fits the training sample.  Deep double descent extends
the phenomenon to neural-network model size, sample size, and training time
\citep{nakkiran2020deep}.  Our work follows the spectral viewpoint but applies
it to the empirical tree-kernel spectrum induced by split candidates.
\citet{curth2023uturn} caution that raw parameter-count axes in tree and
boosting examples can conflate model growth with ensembling or smoothing
effects; this motivates our use of a single split-candidate axis.

\section{Preliminaries}

We consider supervised regression data
\[
  \mathcal{D}_n=\{(x_i,y_i)\}_{i=1}^{n},
  \qquad
  x_i\in\mathcal{X}\subseteq\R^d,\quad y_i\in\R .
\]
Let \(f_t\) denote the ensemble prediction after \(t\) boosting steps and
\[
  r_{t,i}=y_i-f_t(x_i)
\]
the squared-loss residual on the \(i\)-th training point.  A GBDT update fits a
regression tree weak learner \(h_t\) to the residuals and sets
\[
  f_{t+1}(x)=f_t(x)+\varepsilon h_t(x),
\]
where \(\varepsilon>0\) is the learning rate.  The weak learner is a
piecewise-constant decision tree.  Each internal node stores a split test
\(s=(j,b)\) and sends a point left or right according to whether
\(x_j\le b\).  Each leaf stores a constant prediction.  Thus a fitted
tree partitions the input space into leaves and predicts one number on each
leaf.

Modern GBDT implementations make split search finite by binning numerical
features before tree induction.  For feature \(j\), let \(\mathcal{B}_j=\{b_{j,1}<\cdots<b_{j,C_j}\}\) be the selected borders, and let
\(\mathcal C=(\mathcal B_1,\ldots,\mathcal B_d)\) denote the candidate
configuration.  These borders define threshold split candidates
\[
  \begin{aligned}
  \mathcal{S}_{\mathcal C}
  &=
  \{(j,b): b\in\mathcal{B}_j,\ j=1,\ldots,d\},\\
  \mathbf c(\mathcal C)&=(C_1,\ldots,C_d),
  \qquad
  \lvert\mathcal S_{\mathcal C}\rvert=\sum_{j=1}^{d}C_j .
  \end{aligned}
\]

\section{From Split Candidates to Path Features and Tree Kernels}

\begin{figure}[!t]
  \centering
  \includegraphics[width=\columnwidth]{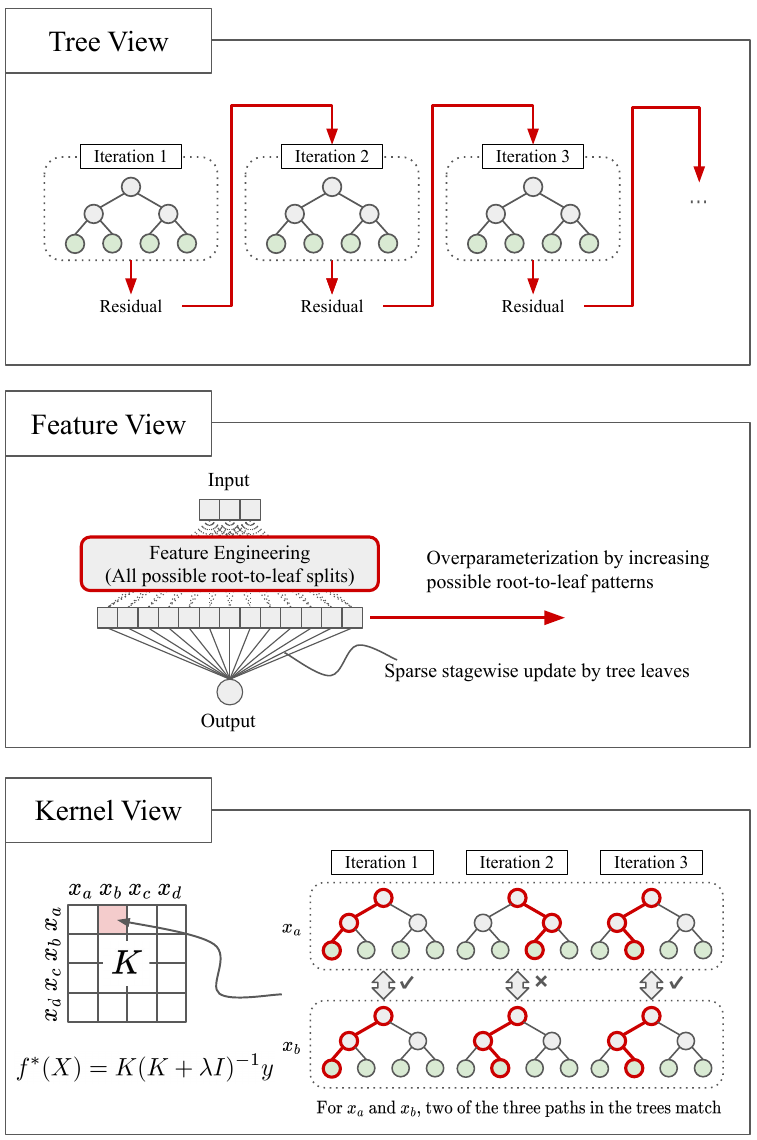}
  \caption{Tree, feature, and kernel views of GBDT training.}
  \label{fig:three-views}
\end{figure}

In this section, we use three linked views of split-candidate scaling to explain
why increasing the number of split candidates can lead to double-descent
behavior.  Figure~\ref{fig:three-views} summarizes the three views.  The
\emph{tree view} starts from leaf-average updates, in which split-candidate count is not yet an
obvious complexity axis.  Holding the loss, tree depth, learning rate, and
number of boosting rounds fixed, the \emph{feature view} reindexes those updates as
sparse moves in a root-to-leaf path dictionary; in this view, increasing split
candidates looks like increasing model width.  The \emph{kernel view} uses the
exact leaf-average identity to build a residual-independent reference kernel,
whose rank and minimum positive eigenvalue diagnose how the candidate-induced
geometry changes with the split-candidate budget.  We do not claim that this reference
kernel exactly describes adaptive GBDT training; its relevance to adaptive
training is tested empirically below.

\subsection{Tree view}
At iteration \(t\), a tree structure \(\nu_t\) partitions the input space into leaves
\(\mathcal{L}_{\nu_t}\).  We write the weak learner \(h_t\) as
\(h_{\nu_t}\) when emphasizing this selected tree structure, and write
\(n_{\nu_t,\ell}(X)=\#\{i:x_i\in\ell\}\) for the number of training points in
leaf \(\ell\).  In the idealized squared-loss update without leaf
regularization, the whole tree update is
\begin{equation}
  h_{\nu_t}(x)
  =
  \sum_{\ell\in\mathcal{L}_{\nu_t}}
  \left(
  \frac{1}{n_{\nu_t,\ell}(X)}
  \sum_{i:x_i\in\ell} r_{t,i}
  \right)\ind\{x\in\ell\},
  \label{eq:tree-view-prediction}
\end{equation}
and boosting updates \(f_{t+1}=f_t+\varepsilon h_{\nu_t}\).
With leaf regularization or second-order objectives, the same leaf-support
representation is retained, with weighted or shrunken leaf values.  Thus the
split-candidate set controls which leaf regions can be selected at each update,
while the number of boosting rounds controls how many updates are taken.

\subsection{Feature view}
The feature view keeps the same leaf regions but changes the indexing.  Fix the
tree depth \(L\) and a split-candidate set \(\mathcal{S}_{\mathcal C}\).  Let
\(\mathcal{P}_{\mathcal C,L}\) be the root-to-leaf path dictionary available to
depth-\(L\) trees under \(\mathcal{S}_{\mathcal C}\).  A path
\(p\in\mathcal{P}_{\mathcal C,L}\) is a conjunction of threshold outcomes and
defines the path feature
\[
  \psi_p(x)=\ind\{x\text{ follows path }p\}.
\]
Boosting selects only a few such features at each step.  If
\(p(\nu_t,\ell)\) denotes the path defining leaf \(\ell\), then one update is
sparse in the ambient path dictionary:
\begin{equation*}
  f_{t+1}(x)
  =
  f_t(x)+\varepsilon
  \sum_{p\in\mathcal{P}_{\mathcal C,L}}
  \Delta\theta_{t,p}\psi_p(x),
\end{equation*}
where
\begin{equation*}
  \Delta\theta_{t,p}
  =
  \begin{cases}
    \displaystyle
    \frac{\sum_{i:x_i\in\ell} r_{t,i}}
         {n_{\nu_t,\ell}(X)},
    &
    p=p(\nu_t,\ell)
    \text{ for some }\ell\in\mathcal{L}_{\nu_t},\\[8pt]
    0,
    &
    \text{otherwise}.
  \end{cases}
\end{equation*}

As illustrated in Figure~\ref{fig:three-views}, split-candidate scaling plays
a role analogous to model width in this feature view.  Increasing the
split-candidate budget enlarges the ambient path dictionary from which boosting
selects its updates, while a depth-\(L\) tree still uses at most
\(|\mathcal{L}_{\nu_t}|\le 2^L\) path features in any one step.  Thus the
number of available update directions grows while each update remains sparse.
As in other feature-based models, expanding this ambient width through the
sample scale creates a natural setting in which double-descent behavior can
arise.

To turn this width analogy into a scale that can be compared with the sample
size, we introduce an ambient dimension proxy for the path-feature space.
Suppose feature \(j\) has \(C_j\) candidate thresholds.  Repeated tests on one
feature remain in its \(C_j\)-dimensional nonconstant threshold span because
their conjunction reduces to an interval or half-line indicator.  A
depth-\(L\) path can combine threshold directions from at most \(L\) distinct
features.  We therefore define the ambient path-span proxy
\[
  D_L(\mathcal C)
  =
  \sum_{\substack{S\subseteq\{1,\ldots,d\}\\ |S|\le L}}
  \prod_{j\in S}C_j ,
\]
where the empty product accounts for the constant direction.  The quantity
\(D_L(\mathcal C)\) is an ambient upper bound on the path-feature dimension,
not a count of paths, trees, or leaves.  On a training sample of size \(n\),
the realized rank is at most \(\min\{n,D_L(\mathcal C)\}\) and can be smaller
when path indicators are zero or linearly dependent.  We therefore use
\(D_L(\mathcal C)\approx n\) only as a reference point for the sample-scale
transition; the kernel view below examines the realized rank and spectrum
directly.

\subsection{Kernel view}
The feature view gives an ambient width proxy.  To study the geometry realized
on the training sample, we express the leaf-average update in
Equation~\eqref{eq:tree-view-prediction} as a kernel operator on the residual
vector.  For \(X=(x_1,\ldots,x_n)\) and a tree \(\nu\), let
\(n_{\nu,\ell}(X)\) be the number of training points in leaf \(\ell\), and let
\(\mathcal{L}_X(\nu)=\{\ell:n_{\nu,\ell}(X)>0\}\).  Following the fixed
tree-kernel construction of \citet{ustimenko2023gradient}, define
\[
  k_{\nu,X}(x,x')
  =
  \sum_{\ell\in\mathcal{L}_X(\nu)}
  \frac{n}{n_{\nu,\ell}(X)}
  \ind\{x\in\ell\}\ind\{x'\in\ell\}.
\]
Leaves empty on \(X\) are omitted.  This kernel is positive semidefinite, with
leaf features
\(\sqrt{n/n_{\nu,\ell}(X)}\ind\{x\in\ell\}\), and its scaling reproduces the
leaf-average update.  If \(z\) lies in an occupied leaf \(\ell\), then
\[
  \frac{1}{n}\sum_i k_{\nu,X}(z,x_i)r_{t,i}
  =
  \frac{1}{n_{\nu,\ell}(X)}
  \sum_{i:x_i\in\ell}r_{t,i}.
\]
Thus
\[
  h_\nu(z)=\frac{1}{n}k_{\nu,X}(z,X)r_t,
\]
which expresses the tree update as a finite-sample kernel expansion over the
training residuals.

To form a reference kernel, we average these single-tree kernels over a
distribution \(\pi_{\mathcal C}\) from which trees are sampled independently of
the residuals:
\[
  \begin{aligned}
  K_{\mathcal C,X}(x,x')
  &=
  \E_{\nu\sim\pi_{\mathcal C}}[k_{\nu,X}(x,x')],\\
  K_{ij}
  &=
  K_{\mathcal C,X}(x_i,x_j)
  \qquad(1\le i,j\le n).
  \end{aligned}
\]
Here \(K_{\mathcal C,X}\) is the averaged kernel function, while \(K\) denotes
its empirical Gram matrix on the training sample.  Because
\(\pi_{\mathcal C}\) does not depend on \(r_t\), \(K\) is fixed, and the
expected tree update is
\[
  \E_{\nu\sim\pi_{\mathcal C}}[h_\nu(z)\mid f_t]
  =
  \frac{1}{n}
  K_{\mathcal C,X}(z,X)(y-f_t(X)).
\]
On the training sample, the corresponding ridge-shrunk update is
\[
  \E[f_{t+1}(X)\mid f_t]
  =f_t(X)+\frac{\varepsilon}{n}
  \{K(y-f_t(X))-\lambda f_t(X)\}.
\]
This is kernel gradient descent (KGD) on the fixed reference kernel.  At
stationarity,
\[
  K(y-f^*(X))-\lambda f^*(X)=0.
\]
Thus, the fitted vector has the kernel ridge regression (KRR) form
\[
  f^*(X)=K(K+\lambda I)^{-1}y,
  \qquad
  \alpha^*=(K+\lambda I)^{-1}y.
\]
Here, the fixed-kernel construction is a residual-independent reference
endpoint, not an assumption that standard GBDTs select trees independently of
the residuals.  Nevertheless, it is connected to ordinary boosting through
the randomized split-selection framework of
\citet{ustimenko2023gradient}.  At finite random strength,
residual-dependent split scores still influence the tree distribution and
induce an \(f_t\)-dependent greedy kernel.  As the random strength increases,
this dependence weakens and the tree distribution approaches a
residual-independent reference distribution.  Fixed-kernel KGD and KRR
therefore provide a controlled setting for isolating candidate-induced
geometry.  The spectral analysis below develops predictions in this reference
setting, and the experiments test whether they persist under
residual-dependent tree construction.

The eigenspectrum of \(K\) provides a finite-sample diagnostic for double
descent.  In the standard spectral view of double descent, test error can peak
as the empirical feature dimension approaches the sample size, especially when
weak regularization makes the fit sensitive to directions with small positive
eigenvalues
\citep{belkin2019reconciling,hastie2022surprises,mei2022generalization}.
For the eigenvalues \(\mu_i(\mathcal C)\) of \(K\), we report
\(\operatorname{rank}(K)=\#\{i:\mu_i>\tau_{\rm eig}\}\) and
\(\mu_{\min}^{+}=\min\{\mu_i:\mu_i>\tau_{\rm eig}\}\), with
\(\tau_{\rm eig}=10^{-10}\) for numerical stability in the eigendecompositions.
These quantities measure the number of directions realized on the training
sample and the strength of the weakest nonzero direction, respectively.

\section{Experiments}

Our experiments test whether split-candidate scaling produces an
intermediate-budget test-error peak associated with empirical rank growth and
small positive kernel eigenvalues.  The experimental design combines
fixed-kernel diagnostics with adaptive tree ensembles to distinguish
candidate-induced geometry from residual-dependent boosting dynamics.

\subsection{Setup}
We summarize the main experimental setup below; full details are provided in Appendix.

\paragraph{Split-candidate controls.}
We use two interfaces exposed directly by CatBoost to control the candidate
set.\footnote{\url{https://catboost.ai/docs/en/references/training-parameters/}}
Throughout, \emph{split-candidate budget} denotes the general capacity control;
\(M\) is the exact total split-candidate count, whereas \(B\) is CatBoost's
native per-feature border budget.
The native interface uses the per-feature \texttt{border\_count} value \(B\)
and lets CatBoost construct the borders, whereas the custom interface uses an
\texttt{input\_borders} file to supply the borders explicitly.  For the latter,
let \(\mathcal C_M\) denote the candidate configuration formed by the first
\(M\) feature-border pairs in a nested ordering; by construction,
\(\lvert\mathcal S_{\mathcal C_M}\rvert=M\).  Both controls enlarge the
split-candidate budget, but \(M\) is the primary axis for testing the
theory because its candidate configurations form monotone prefixes; \(B\)
provides a comparison with CatBoost's native scalar control.

\paragraph{Datasets.}
We use the scikit-learn~\citep{pedregosa2011scikit} California Housing
dataset, derived from the housing data of \citet{pace1997sparse}, as the
primary testbed for the reference-kernel diagnostics, controlled ablations,
training-time analysis, and cross-library comparison.  The dataset contains
\(20{,}640\) rows and \(8\) numerical features.  To assess cross-dataset
robustness, we additionally evaluate \(18\) regression datasets with only
numerical features from
TabularBenchmark~\citep{grinsztajn2022why}; the full dataset list is provided
in Table~\ref{tab:datasets} in Appendix.

\paragraph{Evaluation protocol.}
For the California Housing experiments, unless a panel varies sample size or
depth, we use \(n_{\rm train}=1000\), \(n_{\rm test}=10000\), and tree depth
\(L=2\).  The TabularBenchmark robustness sweep uses
\(n_{\rm train}=1000\), \(n_{\rm test}=5000\), and \(L=2\); the smaller test
split allows the same protocol to cover all \(18\) datasets.  Unless otherwise
stated, we use five resampled training--test splits, a learning rate of \(0.3\),
and up to \(10^6\) CatBoost iterations.  For each repeat, the candidate pool and nested
candidate ordering are rebuilt from the corresponding training split.
Features remain on their original scales because tree split candidates depend
only on within-feature order.  Targets are standardized within each repeat
using the training-split mean and standard deviation, and the same transform
is applied to the test targets.  For a fitted predictor \(f\), we report the
root mean squared error (RMSE).

\paragraph{CatBoost configurations.}
For CatBoost sweeps, we compare two profiles.  The \emph{diagnostic profile},
CatBoost (diagnostic), is designed to approximate the residual-independent
reference-kernel regime.  It uses symmetric trees, plain boosting, zero leaf
\(L_2\) regularization, and \texttt{random\_strength} \(=10^6\); at this scale,
the injected noise dominates residual-dependent differences among split scores,
making split selection effectively random.  It also uses constant model shrinkage at rate
\(\lambda/n_{\rm train}\), with \(\lambda=10^{-5}\).  The \emph{default
profile}, CatBoost (default),
fixes the split borders, depth, learning rate, iteration count, and random seed;
model shrinkage, leaf \(L_2\) regularization, and the remaining boosting choices
are left at CatBoost defaults; split selection therefore remains driven by
residual-dependent scores.

\begin{figure}[t]
  \centering
  \includegraphics[width=\columnwidth]{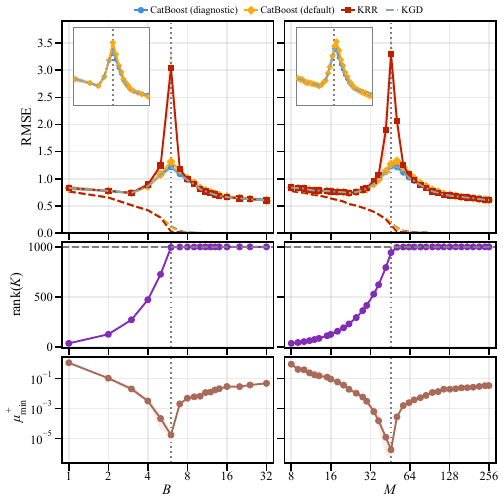}
  \caption{Reference-kernel diagnostics for native border budget \(B\) and
  nested \texttt{input\_borders} size \(M\).  Top-row solid and dashed curves
  denote test and training RMSE, respectively; the dash-dotted KGD reference is
  shown for test only.  Insets zoom in on the CatBoost/KGD test RMSE range, while the
  lower rows show empirical kernel rank and \(\mu_{\min}^+\) of the same
  empirical kernel.  Vertical dotted lines mark the ambient path-span
  crossings.}
  \label{fig:kernel-alignment}
\end{figure}

\begin{figure*}[t]
  \centering
  \includegraphics[width=\textwidth]{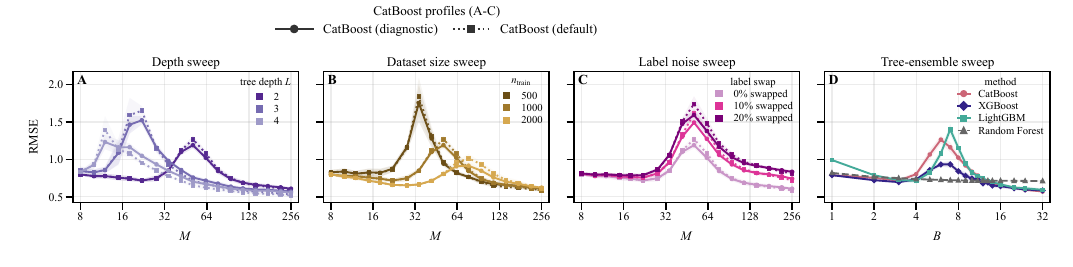}
  \caption{California Housing test RMSE sweeps across depth, sample size, label
  noise, and tree-ensemble algorithm, with panels A--C comparing CatBoost
  profiles; panel D compares three GBDT libraries and a random-forest control~\citep{breiman2001random}
  as \(B\) varies.  Fixed settings are
  \(n_{\rm train}=1000\) for panels A, C, and D and tree depth \(L=2\) for
  panels B--D.}
  \label{fig:catboost-gbdt-sweeps}
\end{figure*}

\subsection{Results}
We report the results in four stages.  We first compare CatBoost with the
fixed-kernel KGD/KRR diagnostics, and then test the predicted depth and
sample-size shifts together with label-noise and cross-library controls.  We
next examine how the peak emerges over training time, and finally assess its
recurrence across TabularBenchmark datasets.

\paragraph{Reference-kernel diagnostics.}
The first experiment tests whether the fixed reference-kernel picture is
visible on the native \(B\) and nested \(M\) axes.
Figure~\ref{fig:kernel-alignment} compares CatBoost (diagnostic), CatBoost
(default), finite-time KGD, and KRR, together with the empirical kernel rank
and \(\mu_{\min}^+\).  On both axes, finite-time KGD closely tracks CatBoost
(diagnostic), with peaks near the ambient path-span crossing; training RMSE is
already close to zero in this region.  CatBoost (default) shows that the peak
remains visible under residual-dependent split selection.  Near the peak,
\(\operatorname{rank}(K)\) approaches \(n\) and \(\mu_{\min}^+\) drops by
several orders of magnitude; on the larger-budget tail, the rank saturates and
\(\mu_{\min}^+\) rises again.  KRR peaks more sharply near the same crossing, as
expected for a stationary ridge fit that uses directions associated with small
positive eigenvalues.  Finite-time KGD and boosting fit these directions only
gradually.  The alignment
of the error peaks with the accompanying rank and eigenvalue changes supports
the proposed spectral account.

\paragraph{Ablations.}
Figure~\ref{fig:catboost-gbdt-sweeps} applies four interventions to California
Housing.  The first two test predicted shifts in peak location, the third tests
whether label noise makes the peak more visible, and the fourth tests both
transfer across GBDT implementations and a non-boosting control.  Panels A--C use the nested
\(M\) sweep and compare the effectively randomized split selection of CatBoost
(diagnostic), which approximates the reference-kernel regime, with the
residual-driven split selection of CatBoost (default).  Panel D instead compares
the native per-feature controls of three GBDT libraries alongside a random-forest
control across \(B\).

Panels A and B test peak location on the nested \texttt{input\_borders} axis
\(M\).  For each nested configuration,
\(\lvert\mathcal S_{\mathcal C_M}\rvert=M\); hence, \(M\) is exactly the total
split-candidate count.  Substituting the balanced-count approximation
\(C_j\approx M/d\) into the path-span proxy gives
\[
  D_L(\mathcal C_M)\approx
  \sum_{r=0}^{L}\binom{d}{r}\left(\frac{M}{d}\right)^r.
\]
The reference crossing is \(D_L(\mathcal C_M)\approx n_{\rm train}\).
Increasing depth adds higher-order path directions and therefore moves the
crossing to smaller \(M\), whereas increasing the training-set size raises the
target dimension and moves it to larger \(M\).  As predicted, deeper trees
peak at smaller \(M\) in Panel A, while larger training sets peak at larger
\(M\) in Panel B.  The reported crossing is the first integer value of \(M\)
satisfying \(D_L(\mathcal C_M)\ge n_{\rm train}\).

Panel C tests peak visibility by increasing the training-label swap rate.
Because label corruption changes neither the candidate configuration nor the
sample size, the proposed mechanism predicts a change in peak prominence
rather than location.  Accordingly, the peak remains in roughly the same range
of split-candidate budgets but becomes more pronounced as the swap rate increases.
This pattern suggests that label noise amplifies the contribution of
variance-sensitive directions near the peak.

Panel D tests cross-library robustness.  XGBoost, LightGBM, and CatBoost all
exhibit an interior peak.  Its recurrence across independently developed
implementations suggests that the phenomenon is not library-specific, but is a
broader feature of GBDTs under split-candidate scaling.  As a
secondary non-boosting control, random-forest test RMSE decreases monotonically
over the evaluated range.  This contrast reflects the difference between direct
forest averaging and inverse-kernel fitting: a forest directly averages leaf
targets, whereas KRR uses
\(\alpha^*=(K+\lambda I)^{-1}y\) and fixed-kernel KGD approaches that solution
through residual iteration.  Small eigenvalues are therefore not
inverse-weighted under forest averaging.  Although this comparison does not
establish causality, it suggests that enlarging the candidate dictionary alone
is insufficient.
Together, these results point to an interaction between candidate-induced
geometry and boosting dynamics.

\begin{figure}[t]
  \centering
  \includegraphics[width=\columnwidth]{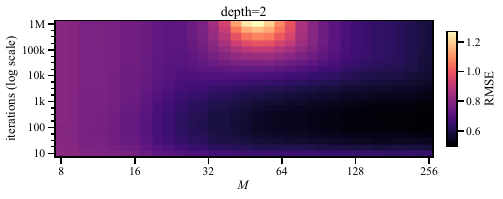}
  \caption{CatBoost (default) test RMSE over split-candidate count \(M\) and
  training checkpoint; color encodes RMSE.}
  \label{fig:fixed-step-heatmap}
\end{figure}

\paragraph{Training-time emergence.}
Figure~\ref{fig:fixed-step-heatmap} tracks the emergence of the
split-candidate peak over training by evaluating fixed CatBoost checkpoints
along the same \(M\) axis.  At small iteration counts, no pronounced
double-descent peak is visible.  As training proceeds, a clear
intermediate-budget peak emerges.  This time dependence accords with the kernel view: unlike closed-form
KRR, which incorporates all nonzero eigendirections, finite-time boosting fits
the variance-sensitive, small-eigenvalue directions only gradually.  This
training-time emergence parallels neural-network studies in which different
components are learned at different rates
\citep{heckel2021early,pezeshki2022multiscale}.  From the
effective-model-complexity viewpoint, early stopping before interpolation can
prevent double descent from becoming visible \citep{nakkiran2020deep}.

\begin{figure*}[t]
  \centering  
  \includegraphics[width=\textwidth]{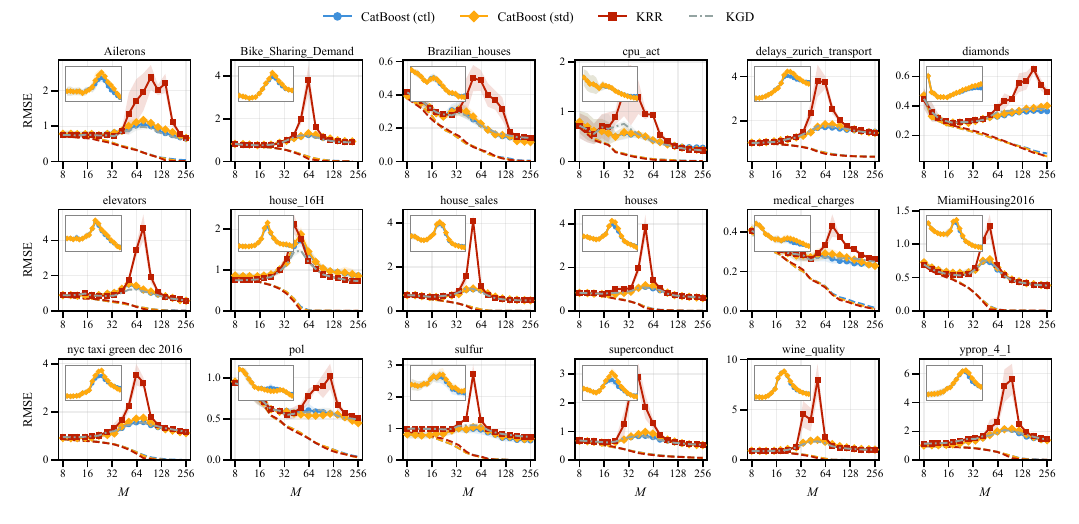}
  
  \caption{TabularBenchmark split-candidate sweeps.  Solid curves show test
  RMSE, dashed curves show training RMSE when reported, and dash-dotted curves
  show the KGD test reference.  Main axes overlay CatBoost (diagnostic),
  CatBoost (default), KRR, and KGD; insets zoom in on the CatBoost test RMSE
  range.}
  \label{fig:tabular-benchmark}
\end{figure*}
\paragraph{Dataset robustness.}
Figure~\ref{fig:tabular-benchmark} reports the robustness sweep over the
\(18\) TabularBenchmark regression datasets with only numerical features,
testing whether double descent under split-candidate scaling extends beyond
the primary California Housing experiments.
KRR exhibits the clearest double-descent peaks across datasets, and the CatBoost profiles show smoother intermediate-budget peaks in
roughly the same ranges of split-candidate budgets.  This pattern agrees with the
proposed mechanism, under which small-eigenvalue directions are expressed
gradually.

We assessed cross-dataset robustness for CatBoost (default) and KRR using a
post-hoc five-fold cross-fit over repeats.  In each fold, four repeats selected
the pair \(M_v<M_p<M_{\max}\) maximizing the mean rise
\(\operatorname{RMSE}(M_p)-\operatorname{RMSE}(M_v)\); the held-out repeat
measured that rise and the subsequent decline to \(M_{\max}\).  Fold
measurements were averaged into one contrast pair per dataset.  Across the
\(18\) datasets, exact one-sided Wilcoxon signed-rank tests assessed each
contrast; the larger \(p\)-value formed the method-level intersection-union
test, followed by Holm correction across the two methods.  Both contrasts were
positive on all \(18\) datasets.  Median rise and decline were \(0.402/0.462\)
for CatBoost (default) and \(1.886/1.879\) for KRR.  The Holm-adjusted
intersection-union \(p\)-value was \(7.63\times10^{-6}\) for each method,
providing statistically significant evidence that both the intermediate peak
and subsequent descent are robust across datasets.

Peak location nevertheless varies because nominal dictionary growth need not
match growth in realized data partitions.  Feature correlations can create
redundant partitions, while uneven bin occupancy creates sparse cells and
finite-sample collisions.  Consequently, empirical rank growth, the spectral
transition, and the error peak are dataset-dependent.  A detailed analysis is
provided in Appendix.

\section{Discussion and Conclusion}

The split-candidate budget is an operational capacity axis for GBDTs: it
expands the root-to-leaf path dictionary while holding depth, ensemble size,
and the leaf-count bound fixed.  Across experiments, intermediate peaks align
with noise-sensitive small-eigenvalue directions.  The monotone random-forest
control shows, however, that dictionary expansion alone is insufficient: its
interaction with the boosting algorithm is essential to the observed double
descent.  Together, the tree, feature, and kernel views trace this capacity
intervention from finer admissible partitions, through an expanding path-feature
dictionary, to the noise-sensitive spectral directions associated with the
intermediate peak.  Peak locations vary with sample size and dataset geometry,
but the pattern recurs across fixed-kernel diagnostics, residual-dependent
boosting, and heterogeneous tabular datasets, distinguishing it from an
implementation-specific artifact.

A long interval separates the initial report of GBDT double descent in
\citet{belkin2019reconciling} from the present demonstration that it can arise
under a single, consistent capacity intervention.  One likely reason is that
grid refinement appears to be a computational setting and leaves tree-level
parameter counts unchanged.  Only in the induced feature/kernel representation
does grid refinement become recognizable as an analogue of parameter scaling:
it expands the dictionary of admissible directions and can increase empirical
rank.  By identifying this previously hidden capacity axis, this paper
demonstrates that double descent can arise in GBDTs through split-candidate
scaling under a single, consistent capacity intervention, with depth, ensemble
size, and the leaf-count bound held fixed.
More broadly, this result shows that an implementation-level control can induce a genuine change in model expressivity. Recognizing such hidden capacity axes may help uncover related generalization transitions in other learning systems.

\clearpage

\bibliography{aaai2027}

\clearpage
\appendix

\section{Details of Experimental Setup}

This section material gives the full experimental protocol and
implementation details for the reported figures, including settings summarized
in the main text.

\subsection{Datasets and Evaluation Protocol}

\paragraph{Datasets.}
California Housing is the primary testbed for reference-kernel diagnostics,
ablations, training-time analysis, and cross-library comparison.  We use the
scikit-learn version~\citep{pedregosa2011scikit}, derived from the data of
\citet{pace1997sparse}, with \(20{,}640\) rows and \(8\) numerical features.
Cross-dataset robustness uses the \(18\) all-numerical regression datasets from
TabularBenchmark~\citep{grinsztajn2022why} in Table~\ref{tab:datasets}.  The
table reports the original size \(n_{\rm full}\), feature count \(d\), and
OpenML~\citep{vanschoren2014openml} identifier for each dataset.

\begin{table}[h]
  \centering
  \caption{Summary of TabularBenchmark~\citep{grinsztajn2022why}
  datasets used in the robustness experiments.}
  \label{tab:datasets}
  \small
  \setlength{\tabcolsep}{1.5pt}
  \begin{tabular}{cccc}
    \toprule
    Dataset & \(n_{\rm full}\) & \(d\) & Source \\
    \midrule
    \texttt{Ailerons} & 13{,}750 & 33 & OpenML 44137 \\
    \texttt{Bike\_Sharing\_Demand} & 17{,}379 & 6 & OpenML 44142 \\
    \texttt{Brazilian\_houses} & 10{,}692 & 8 & OpenML 44141 \\
    \texttt{cpu\_act} & 8{,}192 & 21 & OpenML 44132 \\
    \texttt{delays\_zurich\_transport} & 5{,}465{,}575 & 8 & OpenML 45034 \\
    \texttt{diamonds} & 53{,}940 & 6 & OpenML 44140 \\
    \texttt{elevators} & 16{,}599 & 16 & OpenML 44134 \\
    \texttt{house\_16H} & 22{,}784 & 16 & OpenML 44139 \\
    \texttt{house\_sales} & 21{,}613 & 15 & OpenML 44144 \\
    \texttt{houses} & 20{,}640 & 8 & OpenML 44138 \\
    \texttt{medical\_charges} & 163{,}065 & 3 & OpenML 44146 \\
    \texttt{MiamiHousing2016} & 13{,}932 & 13 & OpenML 44147 \\
    \texttt{nyc-taxi-green-dec-2016} & 581{,}835 & 9 & OpenML 44143 \\
    \texttt{pol} & 15{,}000 & 26 & OpenML 44133 \\
    \texttt{sulfur} & 10{,}081 & 6 & OpenML 44145 \\
    \texttt{superconduct} & 21{,}263 & 79 & OpenML 44148 \\
    \texttt{wine\_quality} & 6{,}497 & 11 & OpenML 44136 \\
    \texttt{yprop\_4\_1} & 8{,}885 & 42 & OpenML 45032 \\
    \bottomrule
  \end{tabular}
\end{table}

\paragraph{Data splits and normalization.}
For each repeat, the experiment seed permutes the rows; the first
\(n_{\rm train}\) form the training set and the next \(n_{\rm test}\) the held-out
test set.  Unless sample size or depth is varied, California Housing uses
\(n_{\rm train}=1000\), \(n_{\rm test}=10000\), and \(L=2\).
TabularBenchmark uses \(n_{\rm train}=1000\), \(n_{\rm test}=5000\), and
\(L=2\), allowing the same protocol for all \(18\) datasets.  Numerical features
remain on their original scales because splits depend only on within-feature
order.  Targets are standardized by the training-target mean and standard
deviation, with the same transformation applied to the test targets.  For a
fitted predictor \(f\), we report root mean squared error (RMSE),
\(\mathrm{RMSE}=\sqrt{n_{\rm test}^{-1}\sum_{i=1}^{n_{\rm test}}(f(x_i)-y_i)^2}\),
where \(y_i\) is the standardized test target.  Reported RMSE values are
therefore in standardized target units.  We use RMSE because it matches the
squared-error training objective and expresses prediction errors on the same
standardized scale across datasets.

\paragraph{Repeats and aggregation.}
Unless otherwise stated, curves average five repeats derived from base seed
\(42\).  Each repeat uses a fresh training--test split and rebuilds its candidate
pool and nested ordering.  Tables and figures report means; shaded bands and
error columns show one standard deviation.  Within each repeat, CatBoost, KRR,
KGD, and the spectral diagnostics share the split and candidate ordering.

\subsection{Split-Candidate Budgets and Construction}

\paragraph{Split-candidate controls.}
We use two CatBoost interfaces to control the candidate set.  The native
interface sets the per-feature \texttt{border\_count} \(B\) and constructs the
borders; the custom interface reads an \texttt{input\_borders} file.  For the
latter, \(\mathcal C_M\) contains the first \(M\) feature-border pairs in a
nested ordering, so \(\lvert\mathcal S_{\mathcal C_M}\rvert=M\).  This exact
\(M\) axis is the primary theory-facing intervention; the native \(B\) axis
provides a comparison with the libraries' per-feature controls.

\paragraph{Sweep grids and checkpoints.}
The experiments use the following fixed grids.
\begin{itemize}
\item The dense nested-\(M\) grid used for the nested-axis reference-kernel,
training-time, and large-sample experiments is 8, 9, 10, 11, 12, 13, 15, 16,
18, 20, 22, 25, 28, 30, 34, 37, 42, 46, 51, 57, 63, 69, 77, 85, 95, 105, 116,
129, 143, 158, 175, 194, 216, 239, and 256.
\item The reduced nested-\(M\) grid used for the controlled depth, sample-size,
and noise sweeps and for TabularBenchmark is 8, 10, 12, 15, 18, 22, 28, 34,
42, 51, 63, 77, 95, 116, 143, 175, 216, and 256.
\item The shared native-\(B\) grid used for the reference-kernel and
tree-ensemble comparisons is 1, 2, 3, 4, 5, 6, 7, 8, 9, 10, 11, 12, 13, 14,
16, 20, 24, and 32.
\item The training-time checkpoints are boosting iteration counts
\(s_k=\operatorname{round}(10^{1+k/4})\), \(k=0,\ldots,20\), which divide
each decade from \(10\) to \(10^6\) into four equal log-scale intervals: 10,
18, 32, 56, 100, 178, 316, 562, 1{,}000, 1{,}778, 3{,}162, 5{,}623, 10{,}000,
17{,}783, 31{,}623, 56{,}234, 100{,}000, 177{,}828, 316{,}228, 562{,}341, and
1{,}000{,}000.
\item The supplementary spectral and KRR analyses use a locally refined
nested-\(M\) grid with values 8, 10, 12, 15,
18, 22, 28, 32, 34, 36, 38, 40, 42, 44, 45, 46, 47, 48, 49, 50, 51, 52, 53,
54, 55, 56, 57, 59, 63, 77, 95, 116, 143, 175, 216, and 256.
\end{itemize}
These grids were fixed before repeat aggregation.  All grid values were
evaluated as sweep points rather than selected as a final operating point by
test performance.

\paragraph{Split-candidate construction.}
For each \texttt{input\_borders} experiment and numerical feature \(j\), let
\(u_{j,1}<\cdots<u_{j,q_j}\) be the sorted unique training values.  We enumerate
\[
  b_{j,k}=\frac{u_{j,k}+u_{j,k+1}}{2},
  \qquad k=1,\ldots,q_j-1,
\]
and retain all such borders.  No training value lies within a consecutive-value
gap, so its midpoint canonically represents the corresponding effective split.
The complete feature-wise lists are fixed before the \(M\) sweep.  We order the
resulting feature-border pairs, write the first \(M\) to a CatBoost
\texttt{input\_borders} file, and thereby obtain nested sets that only grow with
\(M\).  On the native \texttt{border\_count} axis, we train at the specified
\(B\), extract the fitted borders, and reuse them for the external diagnostics.

\paragraph{Round-robin candidate prefixes.}
After enumerating the feature-wise lists, a balanced midpoint-refinement rule
orders each list by selecting the candidate nearest the median rank and
recursing on the left and right sublists.  It creates no new borders; candidates
at each refinement level are seed-permuted.  Each round-robin cycle also
seed-permutes the available features before appending their next candidates.
This preserves broad feature coverage and coarse-to-fine refinement while
randomizing both feature and within-feature order.

\paragraph{Progressive feature activation.}
The TabularBenchmark sweeps in Figure~\ref{fig:tabular-benchmark} use progressive
feature activation so that \(M\) can be smaller than the feature count \(d\).
Each repeat permutes the features.  For \(M<d\), only the first \(M\) are active,
with one effective split each; for \(M\ge d\), all are active and the standard
round-robin prefix is used.  This protocol is limited to the broad robustness
sweeps and associated diagnostics.

\subsection{Reference-Kernel Diagnostics}

\paragraph{External random-tree kernel.}
The random-tree diagnostic uses the same candidate set as the corresponding
CatBoost run but samples tree structures independently of the residuals.  Each
sampled tree is an oblivious tree of the specified depth.  To sample a tree, we
group split candidates by feature.  At each depth, we uniformly select a
feature with unused candidates and then uniformly select an unused candidate
from that feature.  A split candidate is not reused within a tree, whereas
complete tree structures are sampled with replacement.  This procedure
produces a fixed random-tree distribution for
each split-candidate budget, separating candidate-induced geometry from CatBoost's
adaptive split selection.  In the reported diagnostics, we draw \(10^6\)
random-tree structures, matching the maximum number of CatBoost iterations.
Exact duplicates are aggregated before kernel construction, so the empirical
kernel may be constructed from fewer than \(10^6\) unique tree structures.

\paragraph{Kernel normalization, KGD, and KRR.}
For a sampled tree, training points in the same leaf define a normalized leaf
kernel.  Gram-matrix entries for points in the same leaf equal
\(n/n_{\nu,\ell}(X)\), so multiplication by \(K/n\) returns the average residual
in that leaf.  The empirical random-tree kernel is the average of these normalized
leaf kernels over the sampled tree structures.  Kernel ridge regression uses
dual coefficients
\(\alpha^*=(K+\lambda I)^{-1}y\), with \(\lambda=10^{-5}\).  Finite-time
kernel gradient descent uses the same \(\lambda\), learning rate, and iteration
count as CatBoost (diagnostic) under the fixed-kernel update defined in the
kernel view.  For numerical stability, kernel rank and minimum positive
eigenvalue treat eigenvalues at or below \(\tau_{\rm eig}=10^{-10}\) as zero,
preventing floating-point roundoff from being classified as a positive
eigenvalue.  These diagnostics and the KGD/KRR predictions are all computed
from this empirical kernel.

\subsection{Model Profiles and Experimental Sweeps}

\paragraph{Common optimization settings.}
The learning rate is \(0.3\), and the final ensembles use \(10^6\)
boosting iterations unless a training-time checkpoint is being evaluated.
The controlled CatBoost experiments compare the diagnostic and default
profiles described below at the same depth, candidate configuration, iteration
count, learning rate, data split, and random seed.  Unless an experiment
explicitly varies one of these quantities, it uses the default
\(n_{\rm train}=1000\), \(n_{\rm test}=10000\), and \(L=2\) protocol stated
above.

\paragraph{CatBoost profiles.}
CatBoost (diagnostic) is the finite-time CatBoost profile designed to approximate
the residual-independent random-tree diagnostic.  It uses symmetric trees and
plain boosting, with bootstrap, leaf-estimation backtracking, and the initial
average prediction disabled.  We set \texttt{loss\_function} to \texttt{RMSE},
\texttt{score\_function} to \texttt{L2}, \(L_2\) leaf regularization to zero,
\texttt{random\_strength} to \(10^6\), and \texttt{random\_score\_type} to
\texttt{Gumbel}.  At this scale, the
injected noise dominates residual-dependent differences among split scores and
makes split selection effectively random.  The profile also uses model
shrinkage at a constant rate of \(\lambda/n_{\rm train}\), with
\(\lambda=10^{-5}\).  CatBoost (default) fixes only the split-candidate
axis, depth, learning rate, iteration count, and random seed; it does not use
the \(\lambda=10^{-5}\) shrinkage of CatBoost (diagnostic).  Model shrinkage,
leaf \(L_2\) regularization, and CatBoost's other training choices are left at
their library defaults; split selection therefore remains driven by
residual-dependent scores.  The agreement between the external kernel and
CatBoost (diagnostic) supports the proposed mechanism with respect to peak
location and the associated spectral pattern; it does not imply identical
training trajectories.

\paragraph{Controlled depth, sample-size, and noise sweeps.}
The depth panel evaluates \(L\in\{2,3,4\}\) at
\(n_{\rm train}=1000\).  The sample-size panel evaluates
\(n_{\rm train}\in\{500,1000,2000\}\) at \(L=2\).  The label-noise panel
evaluates training-label swap fractions
\(\rho\in\{0,0.1,0.2\}\) at \(n_{\rm train}=1000\) and \(L=2\).
All three panels use \(n_{\rm test}=10000\), both CatBoost profiles, five
resampled splits, and the reduced nested-\(M\) grid.

\paragraph{Label-noise perturbations.}
For the label-noise sweep, corruption is applied only to the standardized
training targets.  Given a swap fraction \(\rho\), we select
\(\mathrm{round}(\rho n_{\rm train})\) training indices without replacement
using the label-noise seed.  The selected targets are then cyclically shifted
within that subset; each corrupted target is replaced by another selected
training target.  Test targets are never corrupted.

\paragraph{TabularBenchmark sweep.}
Each of the \(18\) TabularBenchmark datasets is evaluated with
\(n_{\rm train}=1000\), \(n_{\rm test}=5000\), \(L=2\), both CatBoost
profiles, five resampled splits, the reduced nested-\(M\) grid, and the
progressive feature-activation protocol.  The associated KRR, KGD, and
spectral diagnostics use the same repeat-wise splits and candidate
configurations.

\paragraph{Tree-ensemble comparison protocol.}
The GBDT comparison in Panel D sweeps the native per-feature controls of
CatBoost, XGBoost, and LightGBM.
We set CatBoost's \texttt{border\_count} to \(B\) and the
\texttt{max\_bin} parameter of XGBoost and LightGBM to \(B+1\).  Across the
three libraries, we fix the tree depth at \(L=2\), the ensemble size at
\(10^6\), and the learning rate at \(0.3\), together with the random seed and
threading controls.  All other training parameters remain at their respective
library defaults.  For the
random-forest control, each feature is prequantized into at most \(B+1\)
integer bins using \(B\) evenly spaced effective midpoints between adjacent
unique training values.  Test data are transformed using the same
training-derived borders.  We use \(10^6\) trees of depth \(2\), matching the
GBDT ensemble size and depth.  Beyond these settings, we specify only the
random seed and threading controls; all other random-forest hyperparameters
remain at the scikit-learn defaults.  Thus Panel D aligns the per-feature
split-candidate budget while leaving the split grids and learning algorithms
specific to each method.

\subsection{Software and Hardware}

\paragraph{Libraries.}
All experiments use XGBoost version 3.3.0, LightGBM version 4.6.0, CatBoost
version 1.2.10, and scikit-learn version 1.8.0.

\paragraph{Computational resources.}
All experiments run on a workstation with an Intel Core i9-14900K CPU
(24 cores and 32 threads) and 128 GB of RAM.  The workstation runs Ubuntu 24.04.4
LTS.

\section{Large-Sample Experiment}

\begin{figure}[t]
  \centering
  \includegraphics[width=\columnwidth]{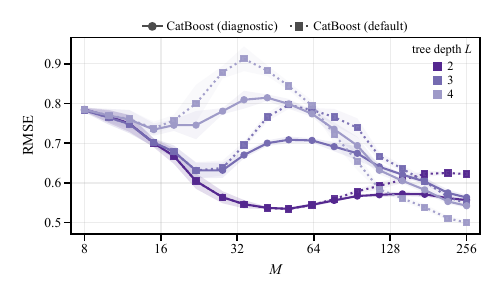}
  \caption{Large-sample California Housing sweep
  (\(n_{\rm train}=10000\)).  Visualization conventions match panel~A of
  Figure~\ref{fig:catboost-gbdt-sweeps}.}
  \label{fig:appendix-california-n10000}
\end{figure}

Figure~\ref{fig:appendix-california-n10000} repeats panel~A of
Figure~\ref{fig:catboost-gbdt-sweeps} at
\(n_{\rm train}=n_{\rm test}=10000\), with all other experimental settings
unchanged.  Double descent remains visible across depths \(2\), \(3\), and
\(4\) and both CatBoost profiles.

\section{Moving Peak Locations}

\begin{figure*}
  \centering
  \includegraphics[width=\textwidth]{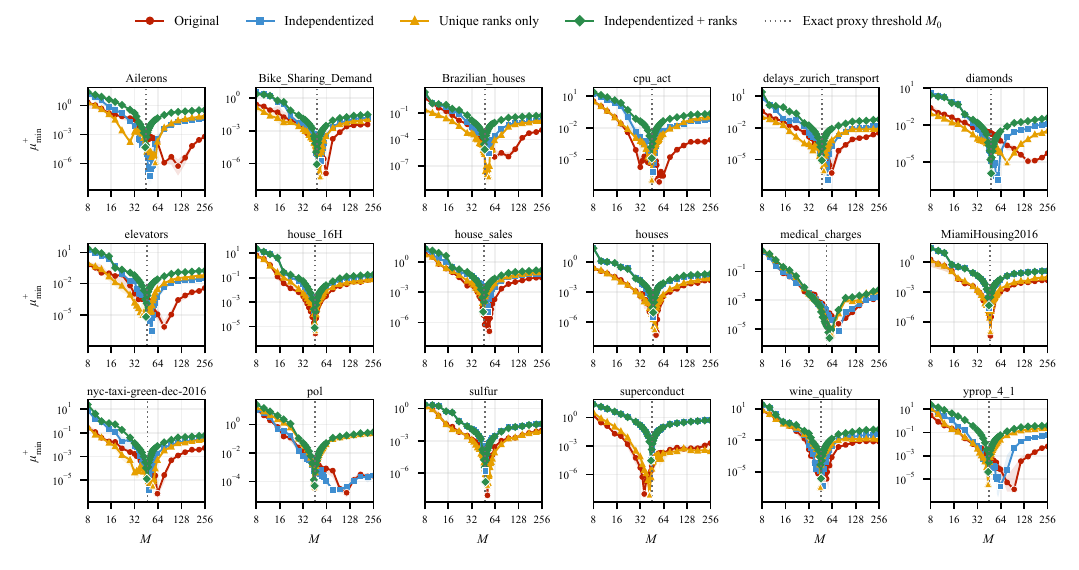}
  \caption{Minimum positive kernel eigenvalue under four data conditions on \(18\)
  datasets.  Curves show medians, and bands show interquartile ranges over five
  repeats.  Enlarged markers denote \(M_\mu\); dotted lines denote
  \(M_0=\min\{M:D_2(\mathcal C_M)\ge n\}\).}
  \label{fig:tabular-geometry-interventions}
\end{figure*}

\begin{figure*}
  \centering
  \includegraphics[width=\textwidth]{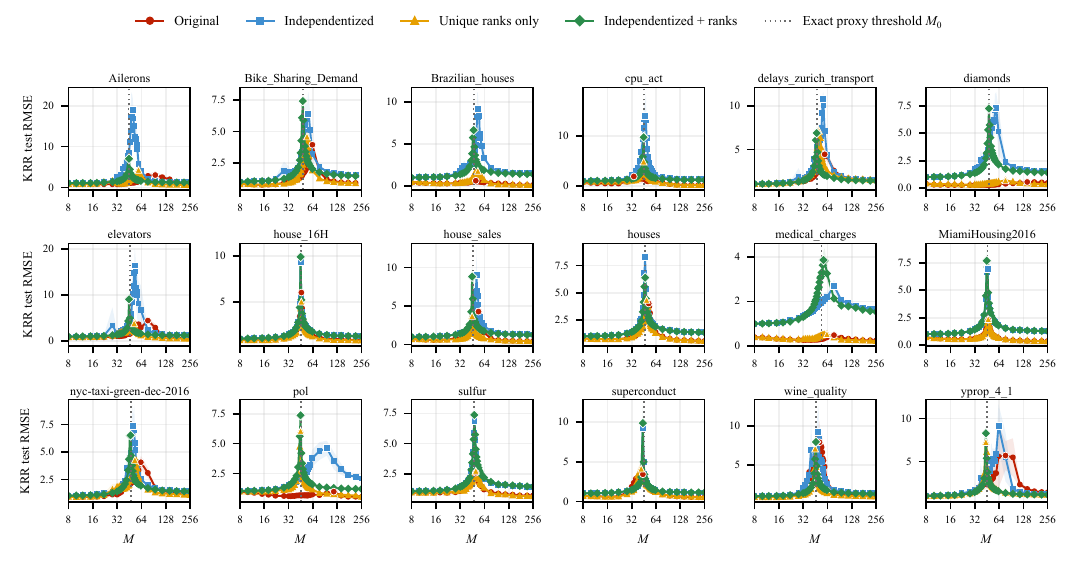}
  \caption{KRR test RMSE for the four data conditions in
  Figure~\ref{fig:tabular-geometry-interventions} (\(\lambda=10^{-5}\)).  Bands
  show one standard deviation over five repeats.  Enlarged markers denote the
  interior test-RMSE maximum; dotted lines denote \(M_0\).}
  \label{fig:tabular-geometry-krr}
\end{figure*}

We test whether two label-free interventions move dataset-specific spectral
transitions toward their geometric crossings by comparing the original data,
each intervention alone, and their combination.

\paragraph{Interventions.}

The \emph{Original} condition uses the observed covariates.  Under
\emph{Independentized}, each feature is permuted separately, removing the
empirical copula while preserving its marginal distribution, effective
thresholds, and feature allocation within the candidate prefix.  This
transformation targets redundant partitions induced by dependent features.
Under \emph{Unique ranks only}, the correspondence of samples across features
is preserved, but each marginal is replaced with unique empirical ranks.  Ties
are broken randomly, and candidates are placed at rank midpoints.  This
transformation balances candidate occupancy, avoids directions associated with
nearly empty leaves, and changes the induced partitions rather than merely
relabeling feature values.  \emph{Independentized + ranks} applies both
transformations.  All four conditions preserve the number of features and use
the same feature-grouped depth-\(2\) tree distribution, including
repeated-feature split pairs.  Seeds and transformations are fixed across
datasets and are not tuned using \(M_\mu\).  Because feature permutation removes
the regression signal, the conditions involving feature permutation are used
to diagnose geometry rather than to compare predictive performance.

\paragraph{Experiments.}

We compare the observed transition
\(M_\mu=\arg\min_M\mu_{\min}^{+}(K_M)\) with the nominal crossing
\(M_0=\min\{M:D_2(\mathcal C_M)\ge n\}\).  The proxy can fail when feature
dependence makes split pairs redundant or occupancy skew creates nearly empty
leaves.  Feature permutation and unique ranking target these effects and
should move \(M_\mu\) toward \(M_0\).
Across the \(90\) dataset--repeat combinations in
Figure~\ref{fig:tabular-geometry-interventions}, the mean absolute
offsets \(\lvert M_\mu-M_0\rvert\) are \(24.6\), \(7.8\), \(4.4\), and
\(0.74\) candidates for \emph{Original}, \emph{Independentized}, \emph{Unique
ranks only}, and \emph{Independentized + ranks}, respectively.  The
combined intervention yields the closest alignment between \(M_\mu\) and
\(M_0\).  Unique ranking has the larger effect for \texttt{pol}, whereas
feature permutation has the larger effect for \texttt{diamonds} and
\texttt{Ailerons},
demonstrating distinct geometric effects.  These spectral curves require
neither labels nor fitted predictors.  The KRR peaks in
Figure~\ref{fig:tabular-geometry-krr} shift similarly: across the same \(90\)
combinations, the mean \(\lvert M_{\rm KRR}-M_0\rvert\) values for the four
conditions are \(21.0\), \(8.2\), \(4.9\), and \(0.67\), respectively, where
\(M_{\rm KRR}\) denotes the interior test-RMSE maximum.

\end{document}